\documentclass[a4paper,english,10pt,twocolumn]{article}
\usepackage[T1]{fontenc}
\usepackage[utf8]{inputenc}
\usepackage{graphicx}
\usepackage[intlimits]{amsmath}
\usepackage{amsfonts, amsthm}
\usepackage{microtype}
\usepackage[english]{babel}
\usepackage{amsfonts, amsthm}
\usepackage{microtype}
\usepackage[left=1.5cm, right=1.5cm, columnsep=.8cm, top=2cm, bottom=2cm]{geometry}
\usepackage[font=footnotesize]{caption}
\usepackage{tikz}
\usepackage{authblk}
\usepackage{algorithmic}

\usepackage[backend=biber,style=numeric,maxnames=3,maxbibnames=9]{biblatex}
\theoremstyle{definition}
\newtheorem{definition}{Definition}

\theoremstyle{plain}
\newtheorem{theorem}[definition]{Theorem}

\newcommand{\RR}{\ensuremath{\mathbb{R}}}

\newcommand{\RP}{\ensuremath{\mathbb{RP}}}
\newcommand{\lapu}{\ensuremath{\mathcal{L}^{\text{up}}}}
\newcommand{\lapd}{\ensuremath{\mathcal{L}^{\text{down}}}}
\newcommand{\lap}{\ensuremath{\mathcal{L}}}
\newcommand{\iso}{\ensuremath{\cong}}

\newcommand{\ie}{}
\def\ie/{i.e.}
\newcommand{\eg}{}
\def\eg/{e.g.}
\newcommand{\etc}{}
\def\etc/{etc.}

\newcommand{\thmcite}[1]{\citeauthor{#1}, \citeyear{#1}~\cite{#1}}

\newcommand{\Reffig}[1]{Figure~\ref{fig:#1}}
\newcommand{\reffig}[1]{figure~\ref{fig:#1}}
\newcommand{\refeq}[1]{equation~\eqref{eq:#1}}
\newcommand{\refsec}[1]{section~\ref{sec:#1}}

\newcommand{\ip}[2]{\ensuremath{\left\langle #1 , #2 \right\rangle}}
\newcommand{\norm}[1]{\ensuremath{\lVert #1 \rVert}}

\DeclareMathOperator{\cut}{cut}
\DeclareMathOperator{\im}{im}
\DeclareMathOperator{\proj}{proj}

\usepackage{color}
\definecolor{darkorchid}{rgb}{0.6,0.196,0.8}

\newcommand{\thmref}[1]{Theorem~\ref{thm:#1}}

\tikzset{twosimp/.style={fill=lightgray}}
\tikzset{twosimpred/.style={fill opacity=0.6,fill=red,draw opacity=0.9}}
\tikzset{threesimp/.style={fill opacity=0.8,fill=blue!60,draw opacity=0.9}}
\tikzset{belowdiag/.style={fill opacity=0.6,fill=gray,color=gray, draw opacity=0.6}}

\definecolor{mplC0}{HTML}{1f77b4}
\definecolor{mplC1}{HTML}{ff7f0e}
\definecolor{mplC2}{HTML}{2ca02c}
\definecolor{mplC3}{HTML}{d62728}
\definecolor{mplC4}{HTML}{9467bd}
\definecolor{mplC5}{HTML}{8c564b}
\definecolor{mplC6}{HTML}{e377c2}
\definecolor{mplC7}{HTML}{7f7f7f}
\definecolor{mplC8}{HTML}{bcbd22}
\definecolor{mplC9}{HTML}{17becf}

\def\figdir{ieee}

\title{A Notion of Harmonic Clustering in Simplicial Complexes}

\author[1]{Stefania Ebli}
\author[1]{Gard Spreemann}

\affil[1]{\footnotesize Laboratory for Topology and Neuroscience,
  École Polytechnique Fédérale de Lausanne, 1015 Lausanne,
  Switzerland}

\date{\today}

\begin{document}
\maketitle

\begin{abstract}
  
  We outline a novel clustering scheme for simplicial complexes that
  produces clusters of simplices in a way that is sensitive to the
  homology of the complex. The method is inspired by, and can be seen
  as a higher-dimensional version of, graph spectral clustering. The algorithm involves only sparse eigenproblems, and is therefore computationally efficient. We believe that it has broad application as a way to extract features
  from simplicial complexes that often arise in topological data
  analysis.
  
\end{abstract}

\section{Introduction}

An important objective in modern machine learning, and part of many scientific and data analysis pipelines, is
\emph{clustering}~\cite{berkhin2007survey}. By clustering, we generally mean the separation of
data into groups, in a way that is somehow meaningful for the domain-specific relationships that govern the underlying data and problem in question. However, the demands that the clustering scheme should satisfy are of course inherently vague.

For data that form a point cloud in Euclidean
space, and where one expects $k$ clusters to exist, one
may employ elementary methods such as
\emph{$k$-means clustering}~\cite{steinhaus1956division}. For data in a more
abstract ``similarity space'', for which no obviously meaningful
Euclidean embedding exists, researchers invented the
schemes~\cite{sorensen1948method, sneath1957application} that we today
refer to as \emph{hierarchical clustering}. Alternatively, one can derive a graph structure from some notion of similarity between the data points. Treating the data points as vertices of a graph allows one to exploit the popular and highly successful spectral clustering techniques~\cite{vonluxburg2007tutorial,ng2001} which developed from the field of spectral graph theory~\cite{chung1997spectral}.

Although the graph structure provides us with additional information about the data,
graphs are intrinsically limited to modeling pairwise interactions. The success of topological methods in studying data, and the parallel establishment of \emph{topological data analysis (TDA)} as a field~\cite{edelsbrunner2000topological, zomorodian2005computing} (see also~\cite{carlsson2008,chazal2017,edelsbrunner2010computational,ghrist2008barcodes} for modern introductions and surveys),
have confirmed the usefulness of viewing data through a higher-dimensional analog of graphs~\cite{moore2012,patania2017}. Such a higher-dimensional analog is called a \emph{simplicial complex}, a mathematical object whose structure can  describe $n$-fold interactions between points. Their ability to capture hidden patterns in the data has led to various applications from biology~\cite{giusti2015,reimann2017} to material science~\cite{hiraoka2016}. Recent work has also expanded classical graph-centric results --- such as Cheeger inequalities~\cite{gundert2014higher,braxton2018}, isoperimetric inequalities~\cite{parzanchevski2016} and spectral methods~\cite{horak2013spectra} --- to simplicial complexes. This leads naturally towards a novel domain of ``spectral TDA'' methods.

In this paper we present the \textit{harmonic clustering algorithm}, a novel clustering scheme inspired by the well-known spectral clustering algorithm for graphs. Our method, like spectral clustering, does not require any parameter optimization and involves only computing the smallest eigenvalue eigenvectors of a sparse matrix. The harmonic clustering algorithm is applied directly to a simplicial complex and it outputs a clustering of the \emph{simplices} (of a fixed degree) that is sensitive to the homological structure of the complex, something that is highly relevant in TDA. Moreover, since simplices can encode interactions of higher order than just the pairs captured by graphs, our algorithm allows us to cluster complex community structures rather than just the entities they comprise.

Our method can be seen as complimentary to the one presented in~\cite{braxton2018}. 

\subsection{Spectral graph theory} \label{sec:specgraph}
The method we present in this paper does not require many formal
results from spectral graph theory. The notions relevant for our purposes are described below for the sake of completeness.

In its simplest form, the \emph{Laplacian} of an undirected and unweighted finite graph $G$ is taken to be the positive definite matrix $L=D-A$, where $A$ is the adjacency matrix of $G$ and $D$ its diagonal degree matrix (\ie/ the row/column sums of $A$). The \emph{normalized Laplacian} is then defined as $\bar{L}=D^{-1/2} L D^{-1/2}$. For reasons that will become clear later on (see \ref{algtop}), we will write $C_0(G)$ for the free real vector space generated by the vertices of $G$, and consider $L$ as the matrix of a linear map $C_0(G)\to C_0(G)$ in this basis.

Already in the middle of the 19th century it was clear that the eigenvalue spectrum of $L$ has a lot to
say about $G$, as is evident from as early as a historic theorem of Kirchhoff relating the eigenvalues of the Laplacian with the number of spanning trees in the graph~\cite{kirchhoff1847}.
From the 1950s, graph theorists and quantum chemists were
independently discovering more relationships between a graph and the
eigenspectrum of its Laplacian. However, the publication of the
book~\cite{cvetkovic1979} may be said to mark the
start of \emph{spectral graph theory} as a field in its own right. A
modern introduction to the field and references to the results listed below can be found in~\cite{chung1997spectral}.

The spectrum of $L$ encodes information about the connectivity of the graph. For instance, the number of connected components of the graph is equal to the dimension of the kernel of $L$. Moreover, the eigenvectors associated to the zero eigenvalues, also called \emph{harmonic representatives}, take constant values on connected components. A perhaps more interesting result is given by the Cheeger constant~\cite{cheeger1969lower}, a measure of how far away a connected graph is from being disconnected by bounding the smallest non-zero eigenvalue of $L$.

\begin{theorem}[\thmcite{cheeger1969lower}; see also \eg/~\cite{chung1997spectral}] Let $G = (V; E)$ be a finite, connected, undirected, unweighted graph. Write $\cut(G)$ for the triples $(S, \bar{S}, \partial S)$ with $S,\bar{S} \subseteq V$ and $\partial S \subseteq E$ such that $S \sqcup \bar{S} = V$ and
\begin{equation*}
  \partial S=\{(u,w)\in E: u\in S, w \in \bar{S}\}. 
\end{equation*}
Define the \emph{Cheeger constant} of $G$ as
\begin{equation*}
  h(G)=\min_{(S,\bar{S},\partial S)\in\cut(G)}\frac{|\partial S|}{\min(\sum_{u\in S}\deg(u),\sum_{w\in \bar{S}} \deg(w))}.
\end{equation*}
Then the first non-zero eigenvalue $\lambda_1$ of the graph's normalized Laplacian satisfies
\begin{equation*}
  2h(G)\geq\lambda_1\geq \frac{(h(G))^2}{2}.
\end{equation*}
\end{theorem}

A partition of $V$ as $S\sqcup \bar{S}$ that attains the Cheeger constant is called a \emph{Cheeger cut}. It is known that finding an exact Cheeger cut is an NP-hard problem~\cite{szlam2010total}. One of the best known approaches to approximating the Cheeger cut is \emph{the spectral clustering method}, which takes the first non-zero eigenvalue eigenvector of the graph Laplacian as a relaxed real-valued solution of the original discrete optimization problem~\cite{vonluxburg2007tutorial}. Namely, the smallest non-zero eigenvector of $\bar{L}$, also called \emph{the Fiedler vector} or \emph{the connectivity vector}~\cite{fiedler1973algebraic}, can be exploited to find the best partition of the graph into two ``almost disconnected'' components. The Cheeger cut can be easily generalized to find $k+1$ ``almost disconnected'' components using the $k$ first non-zero eigenvectors of the graph Laplacian~~\cite{vonluxburg2007tutorial}.

\subsection{Graph spectral clustering}
The Fiedler vector being a relaxed solution of the Cheeger cut has implications for clustering the vertices of a graph into ``almost disconnected'' components~\cite{vonluxburg2007tutorial,belkin2002laplacian}. For the remainder of this section we will assume that the graph under consideration is connected.

\begin{figure}[htbp]
  \vspace*{-0.5cm}
  \centering
  \includegraphics{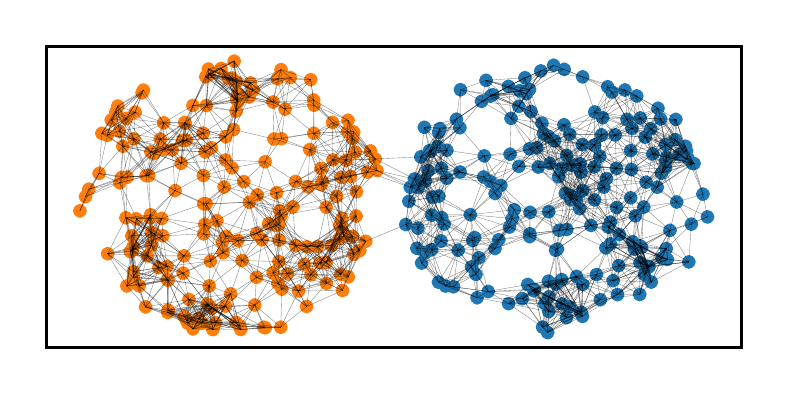}
  \vspace*{-0.5cm}
  \caption{Graph spectral clustering of the nodes of a graph with two well-connected components weakly interconnected. Clustering using the Fiedler vector produces as clusters the well-connected components.} \label{fig:specclust}
\end{figure}

 Graph spectral clustering of a graph $G=(V,E)$ with Laplacian $L$ works in two steps. First, one uses the information encoded in the lowest-eigenvalue eigenvectors of $L$ to map $V$ into low-dimensional Euclidean space. One thereafter uses standard $k$-means or any applicable Euclidean clustering technique on the points in the image of this map, before pulling back to $G$. Specifically, we will write $e_1,e_2,\dotsc,e_n$ for the eigenvectors associated with the $n$ first non-zero eigenvalues of $L$. One defines a function, also called a \emph{spectral embedding}, $\phi:C_0(G)\to\RR^n$ by
\begin{equation}
  \phi(v) = \left(\ip{v}{e_1}, \ip{v}{e_2}, \dotsc, \ip{v}{e_n}\right), \label{eq:graphembed}
\end{equation}
where $\ip{\bullet}{\bullet}$ is the inner product on $C_0(G)$ that makes $V$ orthonormal. As a finite Euclidean point cloud, $\im\phi$ is then clustered in $\RR^n$ by standard $k$-means or any suitable clustering algorithms. The clustering obtained is then pulled back to $V$. \Reffig{specclust} shows an example. Observe that in this case, mapping into the real line using only the Fiedler vector would suffice (\ie/ $n=1$).

As pointed out in \cite{vonluxburg2007tutorial}, spectral clustering is one of the standard approaches to identify groups of ``similar behavior'' in empirical data. It is therefore not surprising that it has been successfully employed in many fields ranging from computer science and statistics to biology and social science. Moreover, compared to other approaches, such as Gaussian Mixture Models clustering, spectral clustering does not require any parameter optimization and can be solved efficiently by standard linear algebra methods.

\section{Harmonic clustering in simplicial complexes}

Our method is inspired by spectral clustering in graphs, but applies instead to a higher-dimensional analog, namely \emph{simplicial complexes}. Instead of clustering only vertices, which are the zero-dimensional building blocks of graphs and simplicial complexes, the method clusters independently building blocks of any dimension.

This section outlines the prerequisite basic constructions from algebraic topology before describing our method. A reader interested in more background on algebraic topology is directed to standard textbooks~\cite{hatcher}.

Those wishing a quick overview of method can view it in algorithmic form in \reffig{alg:harmonic}.

\subsection{Algebraic topology}\label{algtop}

A \emph{simplicial complex} is a collection of finite sets closed under taking subsets. We refer to a set in a simplicial complex as a \emph{simplex} of \emph{dimension $p$} if it has cardinality $p+1$. Such a $p$-simplex has $p+1$ \emph{faces} of dimension $p-1$, namely the sets omitting one element, which we will denote as $(v_0,\dotsc,\hat{v}_i,\dotsc, v_p)$ when omitting the $i$'th element. While this definition is entirely combinatorial, we will soon see that there is a geometric interpretation, and it will make sense to refer to and think of $0$-simplices as \emph{vertices}, $1$-simplices as \emph{edges}, $2$-simplices as \emph{triangles}, $3$-simplices as \emph{tetrahedra}, and so forth.

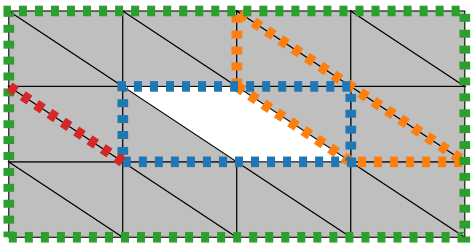
\begin{figure}[htbp]
  \centering
  \begin{tikzpicture}
    \coordinate (v0) at (0, 0);
    \coordinate (v1) at (1.5, 0);
    \coordinate (v2) at (3, 0);
    \coordinate (v3) at (4.5, 0);
    \coordinate (v4) at (6, 0);
    \coordinate (v5) at (0, 1);
    \coordinate (v6) at (1.5, 1);
    \coordinate (v7) at (3, 1);
    \coordinate (v8) at (4.5, 1);
    \coordinate (v9) at (6, 1);

    \coordinate (u0) at (0, 2);
    \coordinate (u1) at (1.5, 2);
    \coordinate (u2) at (3, 2);
    \coordinate (u3) at (4.5, 2);
    \coordinate (u4) at (6, 2);
    \coordinate (u5) at (0, 3);
    \coordinate (u6) at (1.5, 3);
    \coordinate (u7) at (3, 3);
    \coordinate (u8) at (4.5, 3);
    \coordinate (u9) at (6, 3);
    
    \draw[twosimp] (v0) -- (v1) -- (v5) -- cycle;
    \draw[twosimp] (v1) -- (v5) -- (v6) -- cycle;
    \draw[twosimp] (v1) -- (v2) -- (v6) -- cycle;
    \draw[twosimp] (v2) -- (v6) -- (v7) -- cycle;
    \draw[twosimp] (v2) -- (v3) -- (v7) -- cycle;
    \draw[twosimp] (v3) -- (v7) -- (v8) -- cycle;
    \draw[twosimp] (v3) -- (v4) -- (v8) -- cycle;
    \draw[twosimp] (v4) -- (v8) -- (v9) -- cycle;

    \draw[twosimp] (u0) -- (u1) -- (u5) -- cycle;
    \draw[twosimp] (u1) -- (u5) -- (u6) -- cycle;
    \draw[twosimp] (u1) -- (u2) -- (u6) -- cycle;
    \draw[twosimp] (u2) -- (u6) -- (u7) -- cycle;
    \draw[twosimp] (u2) -- (u3) -- (u7) -- cycle;
    \draw[twosimp] (u3) -- (u7) -- (u8) -- cycle;
    \draw[twosimp] (u3) -- (u4) -- (u8) -- cycle;
    \draw[twosimp] (u4) -- (u8) -- (u9) -- cycle;

    \draw[twosimp] (v5) -- (v6) -- (u0) -- cycle;
    \draw[twosimp] (u0) -- (u1) -- (v6) -- cycle;

    \draw[twosimp] (v8) -- (v9) -- (u3) -- cycle;
    \draw[twosimp] (u3) -- (u4) -- (v9) -- cycle;

    \draw[twosimp] (v6) -- (v7) -- (u1) -- cycle;
    \draw[twosimp] (v8) -- (u2) -- (u3) -- cycle;

    \draw[line width=4pt,color=mplC1, dashed, opacity=1] (u2) -- (v8) -- (v9) -- (u3) -- (u7) -- cycle;
    \draw[line width=4pt,color=mplC2, dashed, opacity=1] (v0) -- (v1) -- (v2) -- (v3) -- (v4) -- (v9) -- (u4) -- (u9) -- (u8) -- (u7) -- (u6) -- (u5) -- (u0) -- (v5) -- cycle;
    \draw[line width=4pt,color=mplC0, dashed, opacity=1] (v6) -- (v7) -- (v8) -- (u3) -- (u2) -- (u1) -- cycle;
    \draw[line width=4pt,color=mplC3, dashed, opacity=1] (v6) -- (u0);
    
  \end{tikzpicture}

  \caption{A simplicial complex $K$ with $20$ $0$-simplices, $38$ $1$-simplices (the edges) and $22$ $2$-simplices (the filled triangles), with some highlighted $1$-chains. The highlighted simplices in these represent the edges with non-zero coefficient in each chain (the unfamiliar reader is invited to fill in possible values for these coefficients). The red $1$-chain consists of a single $1$-simplex, and is neither a cycle nor a boundary. The orange $1$-chain has trivial boundary, and is therefore a cycle. It is not a representative of any non-trivial homology class, for it is the boundary of $2$-chain consisting of the three $2$-simplices it encloses. The green and the blue $1$-chains are cycles that represent the same homology class (intuitively the $2$-dimensional hole in the middle). $H_0(K)$ is $1$-dimensional, $K$'s single connected component, while $H_1(K)$ is $1$-dimensional due to the central hole.} \label{fig:bdry}
\end{figure}

Let $C_p(K)$ be the free real vector space with basis $K_p$, the set of $p$-simplices in a simplicial complex $K$. The elements of $C_p(K)$ are called \emph{$p$-chains}. These vector spaces come equipped with \emph{boundary maps}, namely linear maps defined by
\begin{align*}
  &\partial_p:C_p\to C_{p-1} \\
  &\partial_p((v_0,\dotsc,v_p)) = \sum_{i=0}^p (-1)^i(v_0,\dotsc,\hat{v}_i,\dotsc,v_p)
\end{align*}
with the convention that $C_{-1}(K)=0$ and $\partial_0=0$ for convenience. \Reffig{bdry} shows how the boundary maps give a geometric interpretation of simplicial complexes.

One readily verifies that $\partial_p\circ\partial_{p+1}=0$, and so $C_{\bullet}(K)$ is a real \emph{chain complex}. By the \emph{$p$'th homology vector space} of $K$ we will mean the $p$'th homology vector space of this chain complex, namely
\begin{equation*}
  H_p(K)=H_p(C_{\bullet}(K)) = \ker\partial_p/\im\partial_{p+1}.
\end{equation*}
The elements of $\ker\partial_p$ are called \emph{$p$-cycles}, while those of $\im\partial_{p+1}$ are called \emph{$p$-boundaries}, as can be seen geometrically in \Reffig{bdry}. The \emph{Betti numbers} are the dimensions of the homology vector spaces, and we write $\beta_p(K)=\dim H_p(K)$. Intuitively, the Betti numbers count connected components, non-bounding loops, non-bounding cavities, and so forth.

We emphasize again that this is homology with \emph{real} coefficients, not integer or finite field, as is common in TDA.

\subsection{Simplicial Laplacians}

We are in this paper concerned with finite simplicial complexes, and assume that they are built in a way that encodes useful information about the data being studied. We will briefly discuss the case where each simplex in $K$ comes equipped with extra data --- including, but not limited to the filtration/weighting information that is ubiquitous in TDA --- or with a normalization factor derived from the complex's structure, in the form of a function $w:K\to\RR_{+}$. The latter is analogous to the various normalization schemes that are often used in graph spectral theory. Our computational experiments, however, will only consider the case $w=1$.

The weights are encoded into the chain complex by endowing each degree with the inner product that makes all simplices orthogonal, and a simplex have norm given by the weight, \ie/
\begin{align*}
  &\ip{\bullet}{\bullet}_i:C_i(K)\times C_i(K) \to \RR \\
  &\ip{\sigma}{\tau}_i = \begin{cases}
    w(\sigma)^2 &\text{ if } \sigma = \tau\\
    0 &\text{ otherwise.}
    \end{cases}
\end{align*}
Further discussions of weighting schemes can be found in~\cite{horak2013spectra}.

We place no further assumptions on the simplicial complex that we take as input. In particular, it is not necessary for it to come equipped with some embedding into Euclidean space, nor do we demand that it triangulates a Riemannian manifold. Therefore dualities like the Hodge star, which is used to construct the Hodge--de Rham Laplacian in the smooth setting~\cite{madsen1997calculus} that motivates us, are unavailable for our method. The same is true for discrete versions of the Hodge star, such as that of Hirani~\cite{hirani2003thesis}. Instead of dualizing with respect to a Hodge star, to define a discrete version of the Laplacian for simplicial complexes, we simply take the linear adjoint of the boundary operator with respect to the inner product, defining $\partial_i^\ast:C_{i-1}\to C_i$ by
\begin{equation*}
  \ip{\partial_i^\ast \sigma}{\tau}_i = \ip{\sigma}{\partial_i \tau}_{i-1} \quad \forall \sigma\in K_{i-1}, \tau\in K_i.
\end{equation*}
In analogy with Hodge--de Rham theory, we then define the \emph{degree-$i$ simplicial Laplacian} of a simplicial complex $K$ as the linear operator $\lap_i:C_i(K)\to C_i(K)$ such that
\begin{align*}
  &\lap_i = \lapu_i + \lapd_i\\
  &\lapu_i =  \partial_{i+1}\circ\partial_{i+1}^\ast : C_i(K)\to C_i(K) \\
  &\lapd_i = \partial_i^\ast\circ\partial_i : C_i(K)\to C_i(K).
\end{align*}
The \emph{harmonics} are defined as
\begin{equation*}
  \mathcal{H}_i(K) = \ker\lap_i.
\end{equation*}
Observe that there are $p$ Laplacians for a complex of dimension $p$. In most practical applications, the matrices for the Laplacians are very sparse and can easily be computed as a product of sparse boundary matrices and their transposes.

The following discrete version of the important \emph{Hodge decomposition theorem} is a simple exercise in linear algebra in the current setting.
\begin{theorem}[\thmcite{eckmann1944}]\label{thm:eckmann1944}
  The vector spaces of chains decompose orthogonally as 
  \begin{equation*}
    C_i(K) \iso \mathcal{H}_i(K)\oplus\im\partial_{i+1}\oplus (\ker\partial_i)^\perp .
  \end{equation*}  
  Moreover,
  \begin{enumerate}
  \item $\mathcal{H}_i(K)\iso H_i(K)$
  \item the harmonics are both cycles and cocycles (\ie/ cycles with respect to $\partial_{i+1}^\ast$)
  \item the harmonics are the $L^2$-minimal representatives of their (co)homology classes, \ie/ if $h\in\mathcal{H}_i(K)$ and $h\sim z\in\ker\partial_i$ are homologous, then $\ip{h}{h}_i\leq\ip{z}{z}_i$.
  \end{enumerate}
\end{theorem}

The first detailed work on the spectral properties of this kind of simplicial Laplacian was carried out by Horak and Jost~\cite{horak2013spectra}. Recently Steenbergen et al.~\cite{steenbergen2014cheeger} provided a notion of a higher dimensional Cheeger constant for simplicial complexes. At the same time, Gundert and Szedlák~\cite{gundert2014higher} proved a lower bound for a modified version of the higher dimensional Cheeger constant for simplicial complex which was later generalized to weighted complexes by Braxton et al. Mukherjee and Steenbergen~\cite{mukherjee2016random} developed an appropriate notion of random walks on simplicial complexes, and related the asymptotic properties of these walks to the simplicial Laplacians and harmonics. It is worth mentioning that, to the best of our knowledge, no connection between the eigenvectors of the simplicial Laplacian and an optimal cut for simplices in higher dimensions is known.  

Our contribution is a notion of spectral clustering for simplicial complexes using the harmonics.

\subsection{Harmonic clustering}\label{harmclust}

Observe that the ordinary graph Laplacian, as described in \refsec{specgraph}, is just the matrix of $\lap_0=\lapu_0$ in the standard basis for $C_0(G)$. The function $\phi$ in \refeq{graphembed} can thus be seen as projecting the $0$-simplices onto a subspace of low-but-nonzero-eigenvalue eigenvectors of $\lap_0$. The zero part of the spectrum is not used. \thmref{eckmann1944} makes the reason clear: harmonics in $\mathcal{H}_0(G)$ have the same coefficient for every vertex in a connected component of $G$. As connectivity information is easy to obtain anyway, there is little use in adding these eigenvectors to the subspace that $\phi$ projects onto. This is not so for the higher Laplacians. In fact, our method \emph{primarily} uses the harmonics, and only optionally ventures into the non-zero part of the eigenspectrum.

In what follows, $K$ is a fixed simplicial complex arising from data. The particulars of \emph{how} $K$ was built from data is outside the scope of this paper, and is a topic that is well-studied in the field of TDA in general. Our goal is to obtain a useful clustering of $K_p$ for some chosen $p$. We assume that $K$ is of low ``homological complexity'' in degree $p$, by which we mean that $\beta_p(K)$ is small (less than $10$, say).

Analogously to $\phi$ above, we define the \emph{harmonic embedding}
\begin{align*}
  &\psi:K_p\to\RR^{\beta_p(K)} \\
  &\psi = \xi\circ\proj_{\mathcal{H}_p(K)}\circ i,
\end{align*}
where $i:K_p\hookrightarrow C_p(K)$, $\proj:C_p(K)\to\mathcal{H}_p(K)$ is orthogonal projection, and $\xi:\mathcal{H}_p(K)\to\RR^{\beta_p(K)}$ is any vector space isomorphism. In practice, we simply pick an orthonormal basis $h_1,\dotsc,h_{\beta_p(K)}$ for $\mathcal{H}_p(K)$ and let
\begin{equation*}
  \psi(\sigma) = \left(\ip{\sigma}{h_1}_p,\ip{\sigma}{h_2}_p,\dotsc,\ip{\sigma}{h_{\beta_p(K)}}_p\right).
\end{equation*}

In many situations of practical use, it turns out that many points in $\im\psi$ lie along one-dimensional subspaces of $\RR^{\beta_p(K)}$. The membership of a point $\psi(\sigma)$ in such a subspace is what is used to cluster the $p$-simplex $\sigma$ (or to leave it unclustered in case it is not judged to be sufficiently close to lying in one of the subspaces). This amounts to clustering $K_p$ by performing Euclidean subspace clustering of $\im\psi$. A variety of Euclidean subspace clustering methods are available, but are outside the scope of this paper. Examples include \textit{independent component analysis}~\cite{ica}, \textit{SUBCLU}~\cite{subclu}, and density maximization on $\mathbb{S}^{\beta_p(K)-1}$ (or, more precisely, on $\RP^{\beta_p(K)-1}$), which itself has a multitude of approaches, including purely TDA-based ones by means of persistent homology of sublevel sets.

We point out that the choice of the isomorphism $\xi:\mathcal{H}_p(K)\to\RR^{\beta_p(K)}$ does not matter on a theoretical level. It may, however, have practical implications for how easy it is to perform subspace clustering. In experiments we typically choose $\xi$ to be the isomorphism that sends $h_i$ to the standard basis vector $e_i$. Choosing a different orthonormal basis for $\mathcal{H}_p(K)$ then just amounts to an element of $\mathrm{SO}(\beta_p(K))$ acting on $\im\psi$.  

\Reffig{alg:harmonic} summarizes our method in algorithmic form.
\begin{figure}[htbp]
  \begin{algorithmic}
    \REQUIRE Integer $p\geq 0$; simplicial complex $K$ with $\beta_p=\dim(H_p(K))$ small, $K_p=\{\sigma_1,\dotsc,\sigma_N\}$, and inner products $\ip{\bullet}{\bullet}_p$ on $C_p(K)$.
    \STATE $L_p \leftarrow \text{ matrix for } \lap_p$
    \STATE $(h_1,\dotsc, h_{\beta_p})\leftarrow \text{ orthonormal basis for } \ker\lap_p$ (computed using iterative methods~\cite{SLEPc} on $L_p$)
    \FOR{$i=1$ \TO $i=N$}
    \STATE $x_i\leftarrow \left(\ip{\sigma_i}{h_1}_p,\dotsc,\ip{\sigma_i}{h_{\beta_p}}_p\right)$
    \ENDFOR
    \STATE $(a_1,\dotsc,a_k) \leftarrow \mathrm{subcluster}(x_1,\dotsc,x_N)$
    \FOR{$i=1$ \TO $i=k$}
    \STATE $c_i\leftarrow \{\sigma_j\in K_p : j\in a_i\}$
    \ENDFOR    
    \ENSURE Homologically sensitive clustering $c_1,\dotsc,c_k$ of $p$-simplices in $K$.
  \end{algorithmic}
  \caption{Our method in algorithmic form. The subroutine $\mathrm{subcluster}$ refers to any Euclidean subspace clustering scheme, such as independent component analysis~\cite{ica}, SUBCLU~\cite{subclu}, or density maximization on $\mathbb{S}^{\beta_p(K)-1}$ (or, more precisely, on $\RP^{\beta_p(K)-1}$). The latter can be done using methods from TDA, for example by means of persistent homology of certain sublevel sets. Note that there may be unclustered simplices, \ie/ it may happen that $\cup_{i=1}^kc_i \neq K_p$.} \label{fig:alg:harmonic}
\end{figure}

\section{Experimental results} \label{sec:experiments}
In this section we present experimental results for the harmonic clustering algorithm on synthetic data. Specifically, we focus on clustering the edges of various constructed simplicial complexes. The outcomes of our experiments suggest that the harmonic clustering algorithm provides clusters sensitive to the homology of the complex. Comparing our results with those of the traditional spectral clustering algorithm applied to the graph underlying the simplicial complex reinforces the idea that our algorithm reveals substantially different patterns in data compared to the classical method.

Below, we consider four simplicial complexes. Three of them are complexes built from Euclidean point clouds by standard methods from TDA, while one is a triangulation of a torus. We reiterate that our method works with abstract simplicial complexes without utilizing any embedding of these into an ambient space. Euclidean point clouds just happen to be a good and common source of simplicial complexes in TDA, and allow for visualization of the obtained clustering in a way that easily relates to the original data.

An important step in preprocessing many kinds of input data in TDA is constructing a simplicial complex satisfying certain theoretical properties. In particular, if the input data come from points sampled from a topological space $X\subset\RR^n$, one may wish for the homology of the complex to coincide with the homology of $X$.

Two constructions for which some such guarantees exist are the \emph{alpha complex}~\cite{alpha} and the \emph{Vietoris--Rips (VR) complex}~\cite{vietoris1927}. Both can be seen as taking a point cloud and producing a \emph{filtered} simplicial complex $K$, \ie/ a sequence $(K_t)_{t\in\RR_+}$ with the property that $K_s\subset K_t$ whenever $s\leq t$. We wish to work with a single simplicial complex, not a filtration, so we use persistent homology (see \eg/ \cite{ghrist2008barcodes}) to find the filtration scale $t$ for which $K_t$ has the appropriate homology. Of course, since in practice one probably has little or no knowledge of $X$ itself, one cannot necessarily know the ``correct'' $t$ to consider. However, it is often the case in TDA that long-lived homological features --- that is to say, homology classes that remain non-trivial under the induced maps $H_p(K_s)\to H_p(K_t)$ for large $t-s$ --- express interesting properties of the underlying space. We therefore choose a $K_t$ to consider by looking for a scale $t$ within the range of a manageable number of long-lived features and few short-lived ones in the degree under consideration.

In the following experiments, we simplify the setup in the algorithm in \reffig{alg:harmonic} by performing the subroutine $\mathrm{subcluster}$ in a somewhat ad hoc semi-manual way. Specifically, all the images of the $\psi$'s lie in $\RR^2$ or $\RR^3$ in these experiments, so we manually pick out the subspaces $V_1,\dotsc,V_k$ in question. Then, the points in $\im\psi$ are orthogonally projected onto each of the subspaces. A point $\psi(\sigma)$ is determined to lie on subspace $V_i$ if $\proj_{V_i}(\psi(\sigma)/\norm{\psi(\sigma)})$ has norm at least $0.98$, while the onto all other subspaces has norm less than $0.02$. The simplex $\sigma$ is then said to be in cluster number $i$. If the above is not true for any of the subspaces, $\sigma$ is considered unclustered.

In many of the experiments that follow, many points in $\im\psi$ end up determined as ``unclustered'' because they project well onto neither of the $1$-dimensional subspaces, or project too well onto multiple of them, as described in \ref{harmclust}. This is not necessarily a problem, as the parts that are clustered contain a lot of useful information. Moreover, the problem can be reduced by choosing less \textit{ad hoc} subspace clustering methods than we are currently employing.

To ease visualization, we focus on simplicial complexes that naturally live in $\RR^2$ or $\RR^3$ because they arise from point clouds.

\subsection{Wedge of a sphere and two circles} \label{sec:wedge}

\begin{figure}[htbp]
  \centering
  \includegraphics[width=64mm]{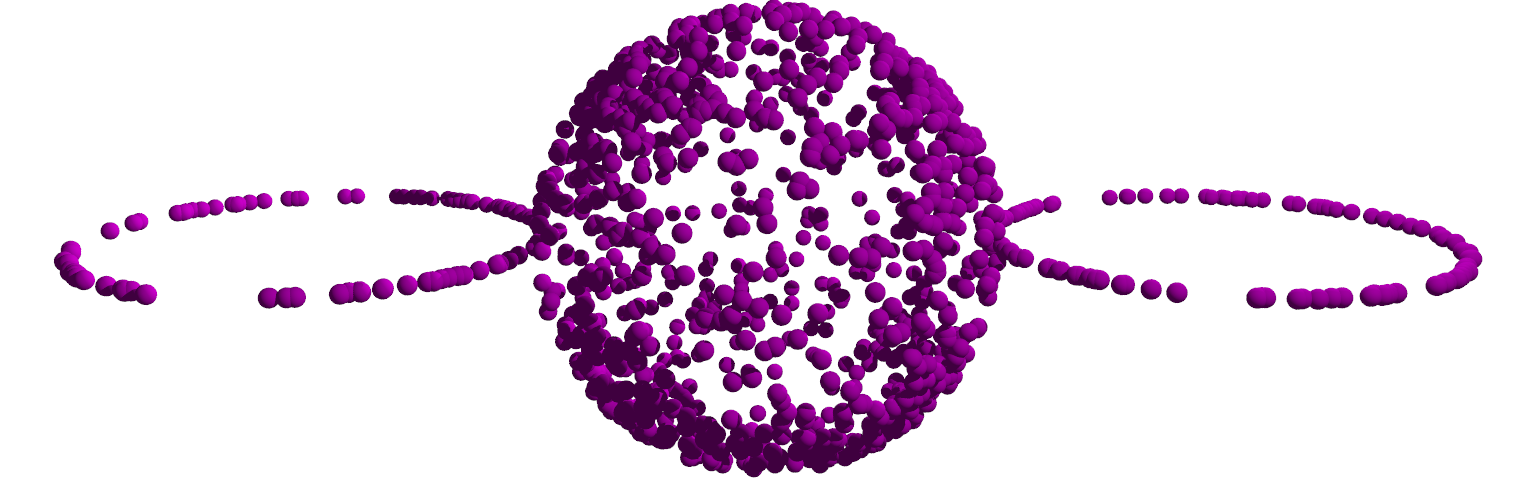}
  \caption{The point sample under consideration in \ref{sec:wedge}.}
\label{fig:pointsmickey}
\end{figure}

In this experiment, we consider a noisy sampling of $X=\mathbb{S}^2\vee\mathbb{S}^1\vee\mathbb{S}^1$ realized as a central unit sphere with unit circles wedged onto antipodal points. We sampled $1000$ points uniformly randomly from the central sphere, adding radial uniform noise with amplitude $0.01$. The circles were sampled using $100$ points each, again with a radial noise of $0.01$. This yields a point cloud with $1200$ points, which is shown in \reffig{pointsmickey}. The VR complex is certainly a suboptimal choice of simplicial complex to build on this kind of data, but we chose it to demonstrate that our method works well also for such an overly dense complex. The complex, constructed at scale $1/2$ and denoted $K$ within this section, has $35722$ $1$-simplices and $485189$ $2$-simplices, and the Betti numbers are $\beta_0(K)=1$, $\beta_1(K)=2$, $\beta_2(K)=1$, as for $X$ itself.

\begin{figure}[htbp]
  \centering
  \includegraphics[width=80mm]{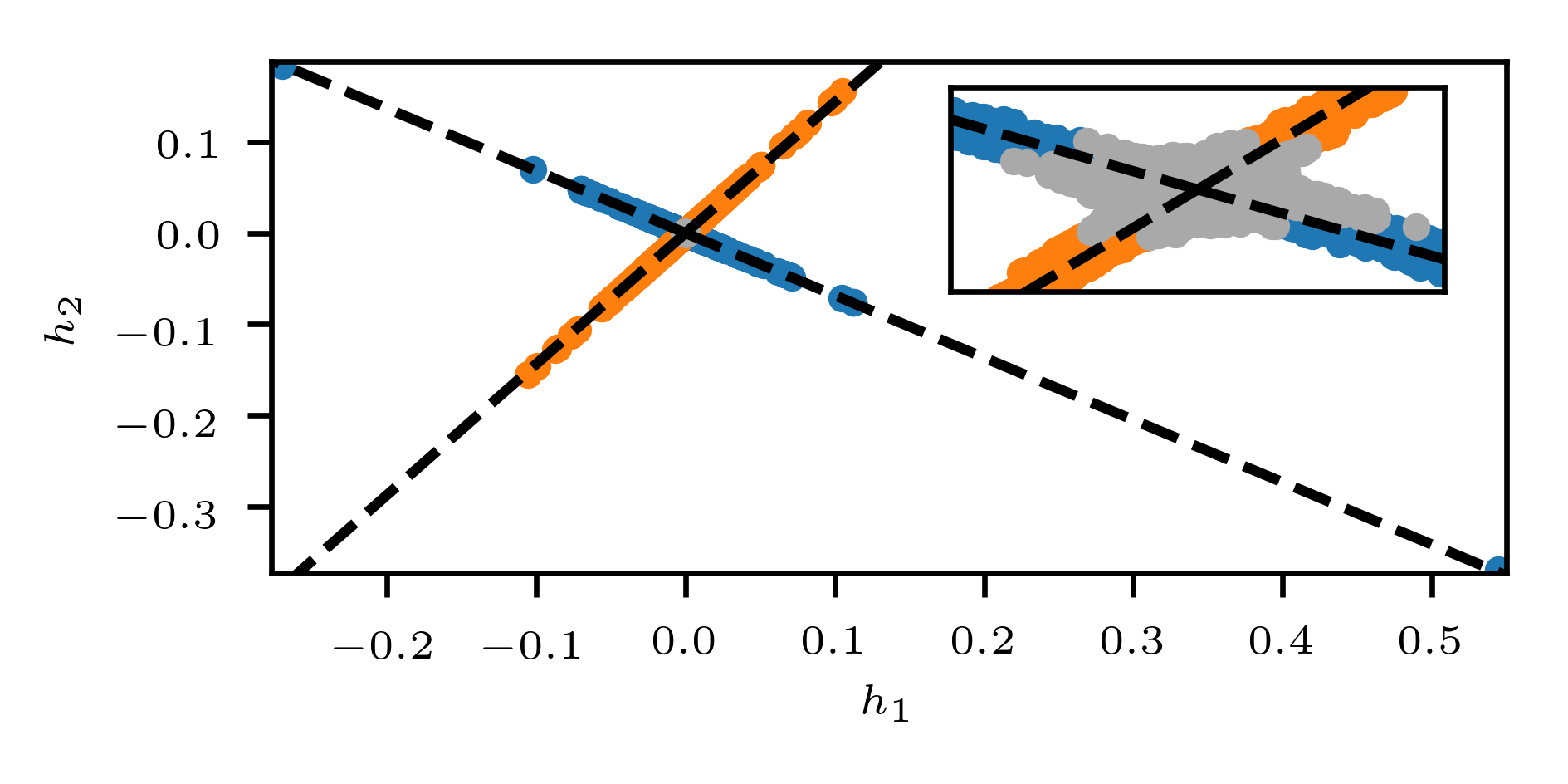}
  \vspace*{-0.5cm}
  \caption{The image of $\psi$ for the $1$-simplices in the VR complex from the experiment in \ref{sec:wedge}. The dashed lines indicate the subspaces used for clustering. The inset shows a detailed view near the origin, where one can see a large number of points in gray that are unclustered due to them projecting too well onto both subspaces.}
  \label{fig:wedge-embed}
\end{figure}

\begin{figure}[htbp]
  \centering
  \includegraphics[width=80mm]{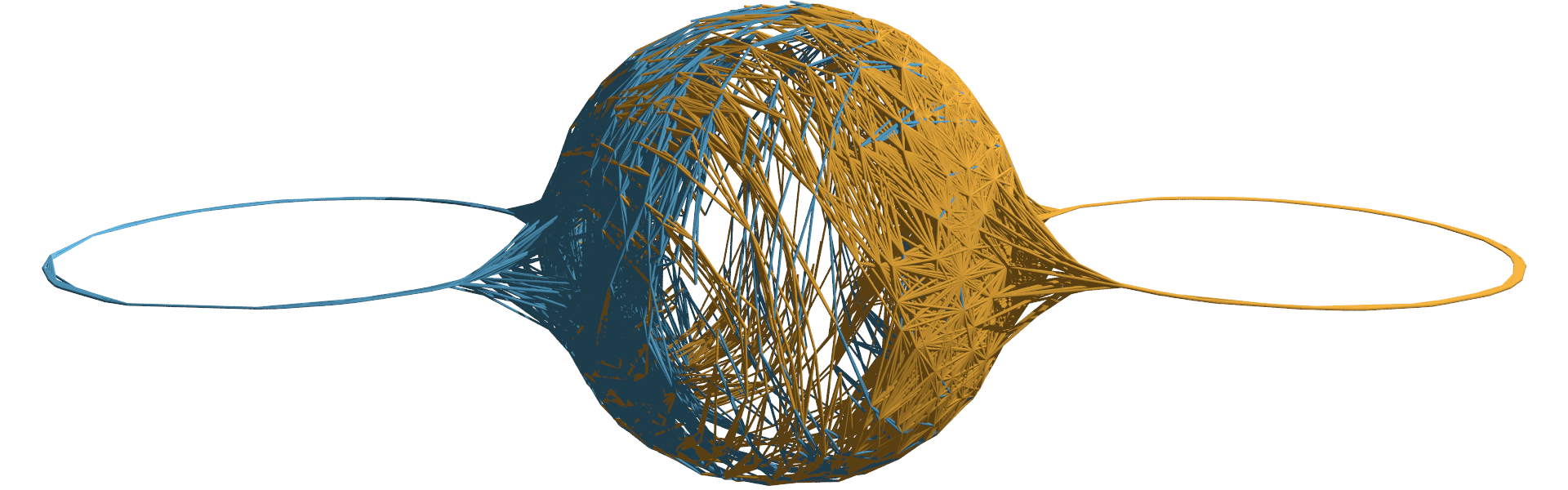}
  \caption{The clustering from \reffig{wedge-embed} pulled back to the $1$-simplices of the VR complex from \ref{sec:wedge}, which is here drawn in $\RR^3$ using the coordinates of the points for visualization purposes only. The unclustered $1$-simplices, $19254$ in number, are not drawn.}
  \label{fig:wedge-clustered}
\end{figure}

We focus on clustering the $1$-simplices of the complex. The image of $\psi$ in $\RR^2$ is shown in \reffig{wedge-embed}. The points are colored according to which of the two one-dimensional subspaces they are deemed to belong to. The determination was made by a simple criterion of projecting well enough onto one of the lines, but not the other. Points that project well onto both or neither are considered unclustered and shown as red. \Reffig{wedge-clustered} shows this clustering pulled back to the complex itself, excluding the unclustered edges. Observe how the method separates the $1$-simplices of the VR complex in a manner that is sensitive to the two non-bounding cycles that generate homology in degree $1$ (the two circles).

\begin{figure}[htbp]
  \centering
  \includegraphics[width=80mm]{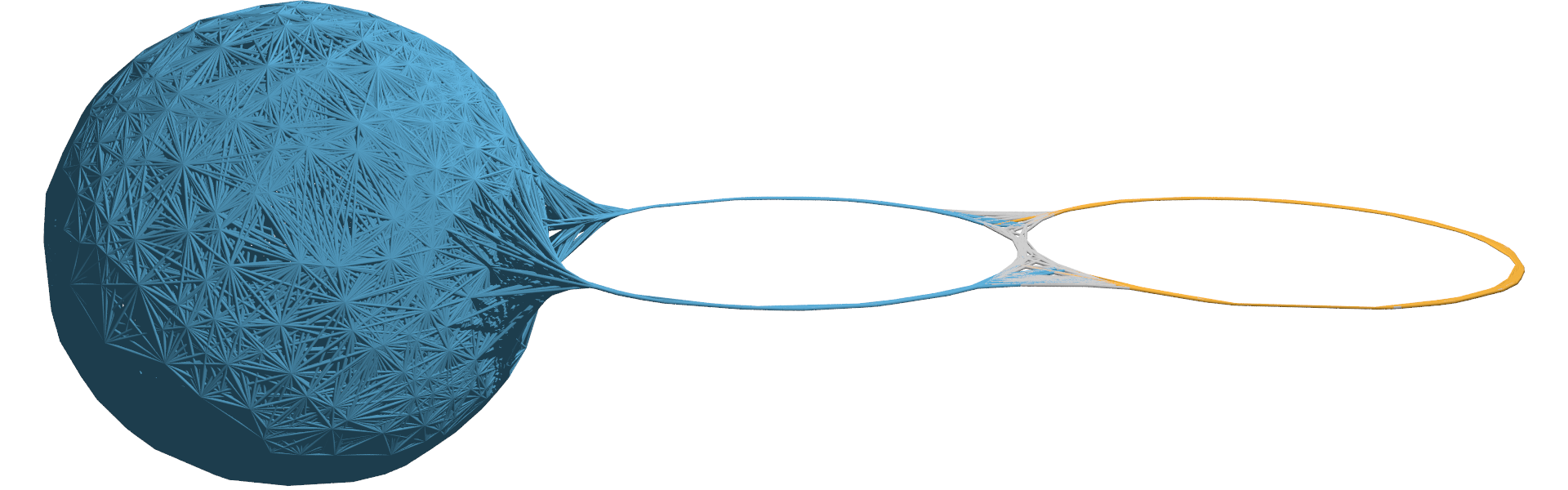}
  \caption{The result of clustering the rearranged point cloud from \ref{sec:wedge}. Again the $1$-simplices of the VR complex are clustered in a way respecting the generators of 1-homology. The unclustered simplices, $467$ in number, are drawn in gray. (That the sphere appears solid is only a visualization artifact; the $2$-simplices are not drawn.)} 
  \label{fig:retarded-wedge-clustered}
\end{figure}

We also repeated the experiment with one of the circles in $X$ moved to be attached to the other circle instead of the sphere. This space is obviously homotopy equivalent to $X$, but is geometrically very different. \reffig{retarded-wedge-clustered} shows the result. Observe that the sphere is now captured by its adjacent circle, and that the unclustered edges tend to be those near where the two circles intersect.

\subsection{Punctured plane} \label{sec:punctured}

\begin{figure}[htbp]
  \centering
  \includegraphics{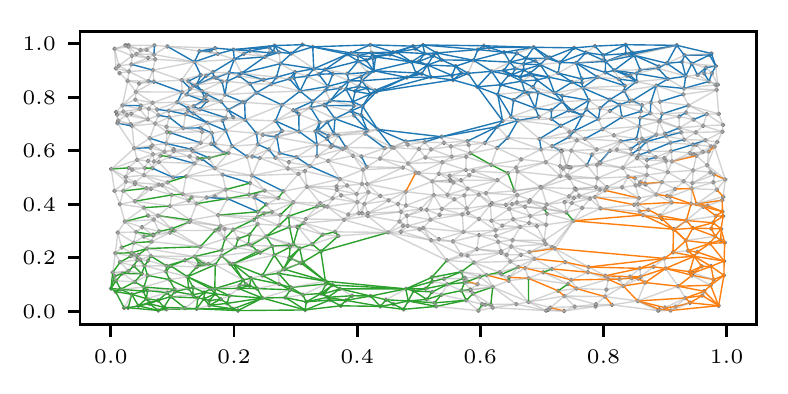}
  \vspace*{-0.5cm}
  \caption{The point cloud of the experiment in \ref{sec:punctured} is shown as faint dots. The punctures can be seen in near $(0.5, 0.8)$, $(0.4, 0.2)$ and $(0.8, 0.3)$, and one observes that the clustered $1$-simplices (blue, green, orange, respectively) follow the punctures. The gray $1$-simplices are unclustered. The $2$-simplices have not been drawn.} \label{fig:planeclust}  
\end{figure}

In this experiment we uniformly randomly sample $1000$ points from a unit square in $\RR^2$ with three disks of radius $1/10$ cut out. The points are seen as faint does in \reffig{planeclust}. We construct the alpha complex at parameter $0.1$, and denote it by $K$ in this section. It has Betti numbers $\beta_0(K)=1$, $\beta_1(K)=3$ and $\beta_{i>1}(K)=0$. There are $2914$ $1$-simplices and $1912$ $2$-simplices. We again focus on the $1$-simplices for clustering. The codomain of $\psi$ is now $\RR^3$, and the image is clustered according to three $1$-dimensional linear subspaces. The result is shown in \reffig{planeclust}, and we again observe how the obtained clustering occurs with respect to the punctures of the square.
 
\subsection{Torus}

We next perform clustering of the edges of two different tori.

\subsubsection{From a point cloud} \label{cloudtorus}

We uniformly randomly sampled $1500$ points from the unit square and map these under $(\varphi,\theta)\mapsto\left((2 + \sin(2\pi\varphi))\cos(2\pi\theta), (2 + \sin(2\pi\varphi))\sin(2\pi\theta), \cos(2\pi\varphi)\right)$ to produce a point sample of a torus in $\RR^3$. The points were then given a uniformly random noise of amplitude $0.01$ in both radii. Again a VR complex $K$ was built, at scale $0.8$. It has $35270$ $1$-simplices and $377873$ $2$-simplices, and has the homology of a torus, \ie/ $\beta_0(K)=1$, $\beta_1(K)=2$, $\beta_2(K)=1$. VR was chosen in order for the clustering task to be more complicated than in a more orderly alpha complex.

\begin{figure}[htbp]
  \centering
  \includegraphics[width=80mm]{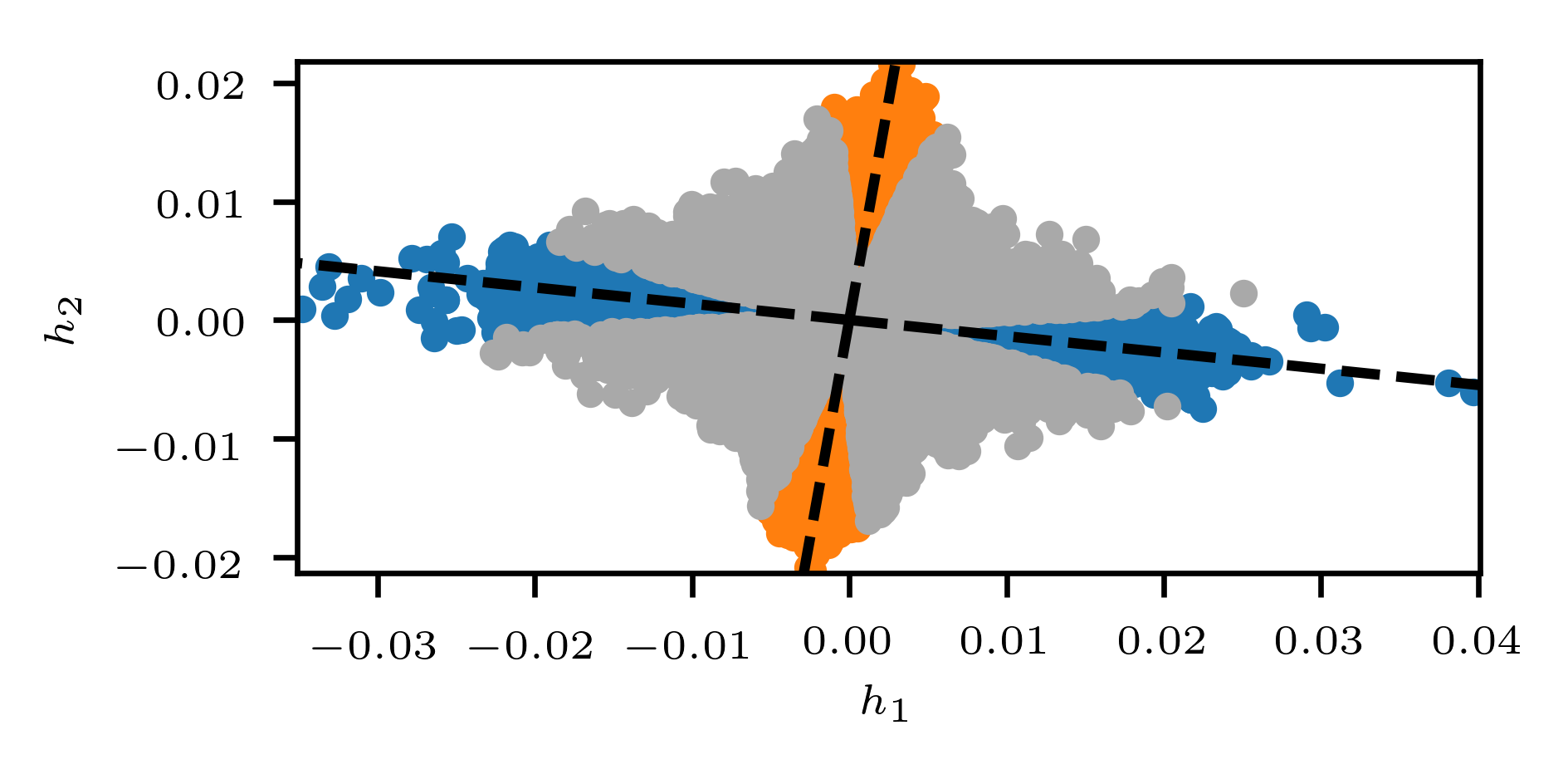}
  \vspace*{-0.5cm}
  \caption{The clustering of the $1$-simplices from the simplicial complex obtained from the sampled torus in the experiment in \ref{cloudtorus}. The unclustered points are shown in gray. They are $23103$ in number.} \label{fig:torus-embed}
\end{figure}

\begin{figure}[htbp]
  \centering
  \includegraphics[width=80mm]{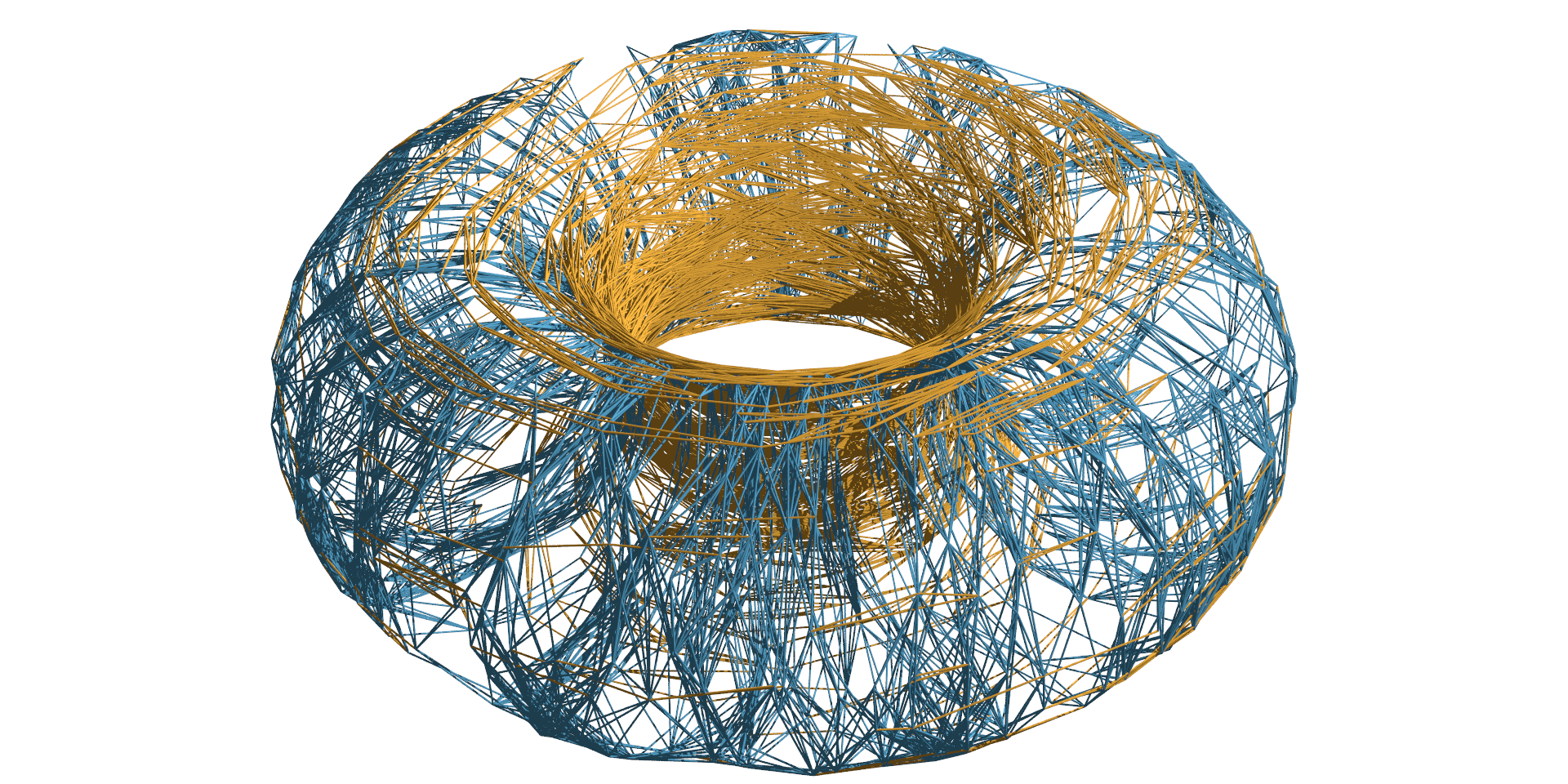}
  \caption{The clustering in \reffig{torus-embed} pulled back to $1$-simplices of the torus from the experiment in \ref{cloudtorus}. The unclustered ones are not shown, something which may make the torus appear broken.} \label{fig:torus-clustered}
\end{figure}

\Reffig{torus-embed} shows the image in $\RR^2$ of $K_1$ under $\psi$. The subspaces for clustering are somewhat harder to make out than before, but they can still be found. The result of clustering by them can be seen in \reffig{torus-clustered}. Observe that the two clusters respect the two independent unfilled loops of the torus.

\subsubsection{A triangulation of the flat torus} \label{sec:flattorus}

\begin{figure}[htbp]
  \centering
  \includegraphics{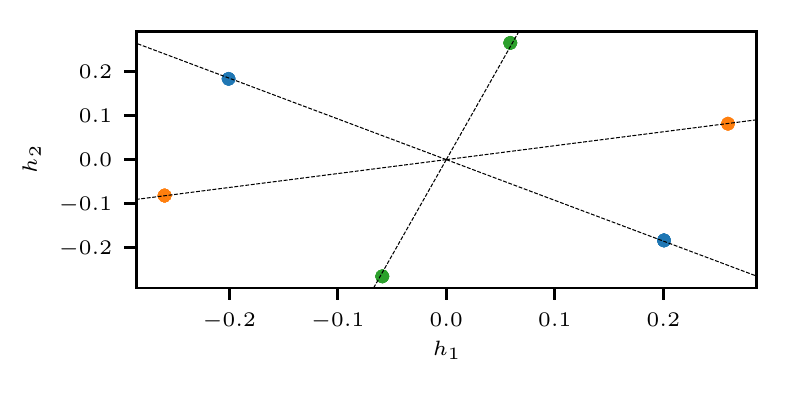}
  \vspace*{-0.5cm}
  \caption{The clustering of the $1$-simplices from the simplicial complex obtained as a triangulation of the flat torus from the experiment in~\ref{sec:flattorus}. Note that many points overlap. Three clusters are given by points lying on three different linear subspaces.}
\label{fig:flattorus-embed}  
\end{figure}

\begin{figure}[htbp]
  \centering
  \begin{tikzpicture}
    \coordinate (a0) at (0,0);
    \coordinate (b0) at (1.5,0);
    \coordinate (c0) at (3,0);
    \coordinate (d0) at (4.5,0);

    \coordinate (a1) at (0,0.7);
    \coordinate (b1) at (1.5,0.7);
    \coordinate (c1) at (3,0.7);
    \coordinate (d1) at (4.5,0.7);

    \coordinate (a2) at (0,1.4);
    \coordinate (b2) at (1.5,1.4);
    \coordinate (c2) at (3,1.4);
    \coordinate (d2) at (4.5,1.4);

    \coordinate (a3) at (0,2.1);
    \coordinate (b3) at (1.5,2.1);
    \coordinate (c3) at (3,2.1);
    \coordinate (d3) at (4.5,2.1);
    
    \draw[twosimp] (a0) -- (b0) -- (a1) -- cycle;
    \draw[twosimp] (b0) -- (b1) -- (a1) -- cycle;

    \draw[twosimp] (b0) -- (c0) -- (b1) -- cycle;
    \draw[twosimp] (c0) -- (c1) -- (b1) -- cycle;

    \draw[twosimp] (c0) -- (d0) -- (c1) -- cycle;
    \draw[twosimp] (d0) -- (d1) -- (c1) -- cycle;

    \draw[twosimp] (a1) -- (b1) -- (a2) -- cycle;
    \draw[twosimp] (b1) -- (b2) -- (a2) -- cycle;

    \draw[twosimp] (b1) -- (c1) -- (b2) -- cycle;
    \draw[twosimp] (c1) -- (c2) -- (b2) -- cycle;

    \draw[twosimp] (c1) -- (d1) -- (c2) -- cycle;
    \draw[twosimp] (d1) -- (d2) -- (c2) -- cycle;

    \draw[twosimp] (a2) -- (b2) -- (a3) -- cycle;
    \draw[twosimp] (b2) -- (b3) -- (a3) -- cycle;

    \draw[twosimp] (b2) -- (c2) -- (b3) -- cycle;
    \draw[twosimp] (c2) -- (c3) -- (b3) -- cycle;

    \draw[twosimp] (c2) -- (d2) -- (c3) -- cycle;
    \draw[twosimp] (d2) -- (d3) -- (c3) -- cycle;

    \draw[color=mplC0, line width=2pt] (a0) -- (d0);
    \draw[color=mplC0, line width=2pt] (a1) -- (d1);
    \draw[color=mplC0, line width=2pt] (a2) -- (d2);
    \draw[color=mplC0, line width=2pt] (a3) -- (d3);

    \draw[color=mplC1, line width=2pt] (a0) -- (a3);
    \draw[color=mplC1, line width=2pt] (b0) -- (b3);
    \draw[color=mplC1, line width=2pt] (c0) -- (c3);
    \draw[color=mplC1, line width=2pt] (d0) -- (d3);

    \draw[color=mplC2, line width=2pt] (b0) -- (a1);
    \draw[color=mplC2, line width=2pt] (c0) -- (a2);
    \draw[color=mplC2, line width=2pt] (d0) -- (a3);
    \draw[color=mplC2, line width=2pt] (d1) -- (b3);
    \draw[color=mplC2, line width=2pt] (d2) -- (c3);
  \end{tikzpicture}

  \caption{A triangulation of the flat torus represented as a rectangle with pairs of opposing edges identified. The $1$-simpilces are clustered into three groups (orange, blue and green). Those in orange and blue are representatives of the two $1$-homology classes of the complex, whereas the green ones are a linear combination of the others. } \label{fig:flattorus-clustered} 
\end{figure}
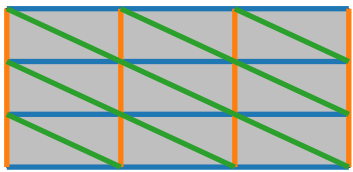

As a smaller, more abstract and noise-free example, we consider a triangulation of a flat torus. The considered triangulation consists of a simplicial complex with $9$ vertices, $27$ $1$-simplices and $18$ $2$-simplices. The image of its $1$-simplices in $\RR^2$ under $\psi$ is shown in \reffig{flattorus-embed}. The arrangement into a perfect hexagon means that there are in fact \emph{three} subspaces that can be chosen for clustering. The clusters are shown in \reffig{flattorus-clustered}. The arrangement into a hexagon, and therefore the result of three instead of the expected two clusters,  disappear if one breaks some of the symmetry in the triangulation, for example by having some of the diagonal edges go the opposite direction.

\subsection{Clustering $2$-simplices} \label{sec:twospheresclust}

\begin{figure}[htbp]
  \centering
  \includegraphics[width=60mm]{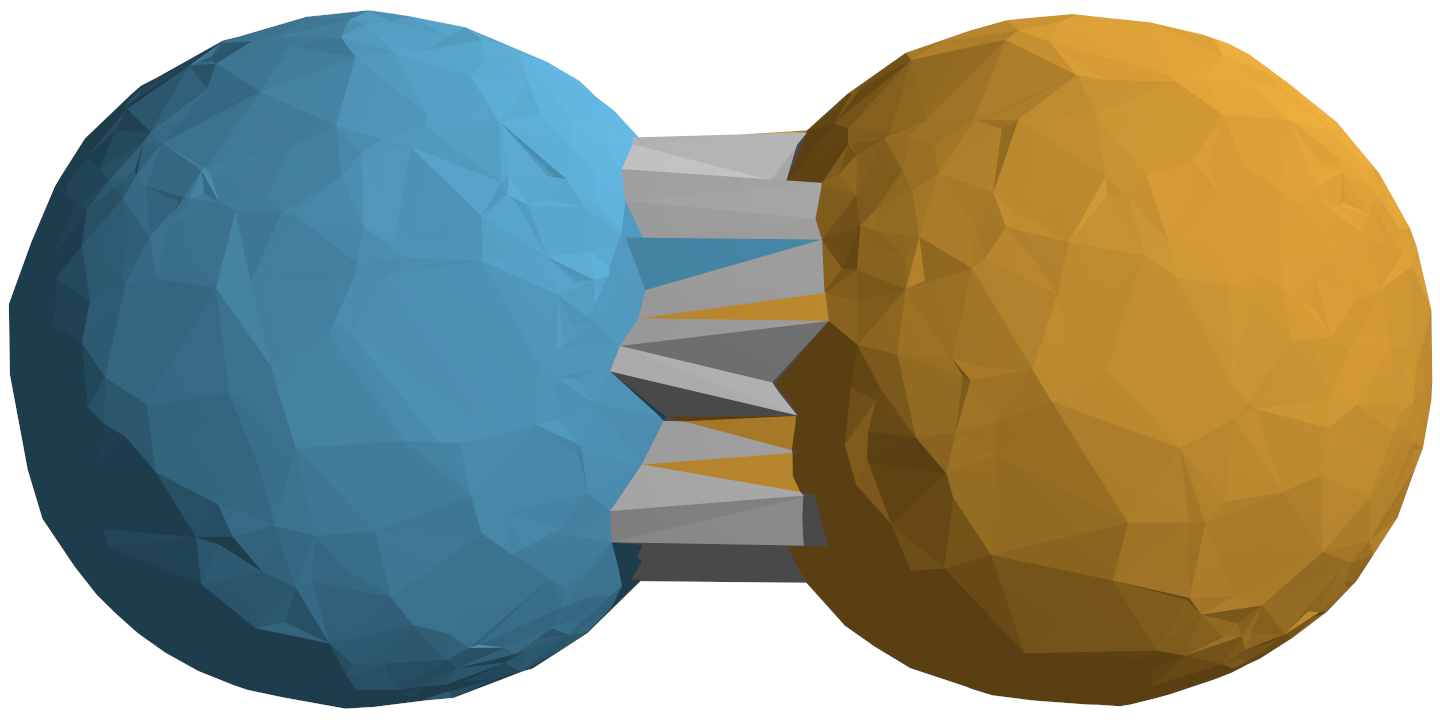}
  \caption{The output of our method when clustering the $2$-simplices from the complex in the experiment in~\ref{sec:twospheresclust} are the blue and orange clusters. The $1485$ gray simplices are unclustered.} \label{fig:twospheresclust}  
\end{figure}

We have illustrated our method only on $1$-simplices so far for ease of visualization. To point out that it also performs well in other dimensions, we sampled $1000$ points (each) from two spheres of radius $1$ centered at $(-1,0,0)$ and $(1,0,0)$, each with a radial uniform random noise with amplitude $0.01$. We computed the alpha complex $K$ at parameter $0.3$, so as to create a rather messy region between the spheres. There are $8851$ $1$-simplices and $10478$ $2$-simplices, and $\beta_0(K)=1$, $\beta_1(K)=0$ and $\beta_2(K)=2$ as expected. Our clustering method performs as expected, producing clusters of $K_2$ that correspond to homological features, as is shown in \reffig{twospheresclust}.

\subsection{Comparison with graph spectral clustering}

\begin{figure}[htbp]
  \centering
  \includegraphics[width=64mm]{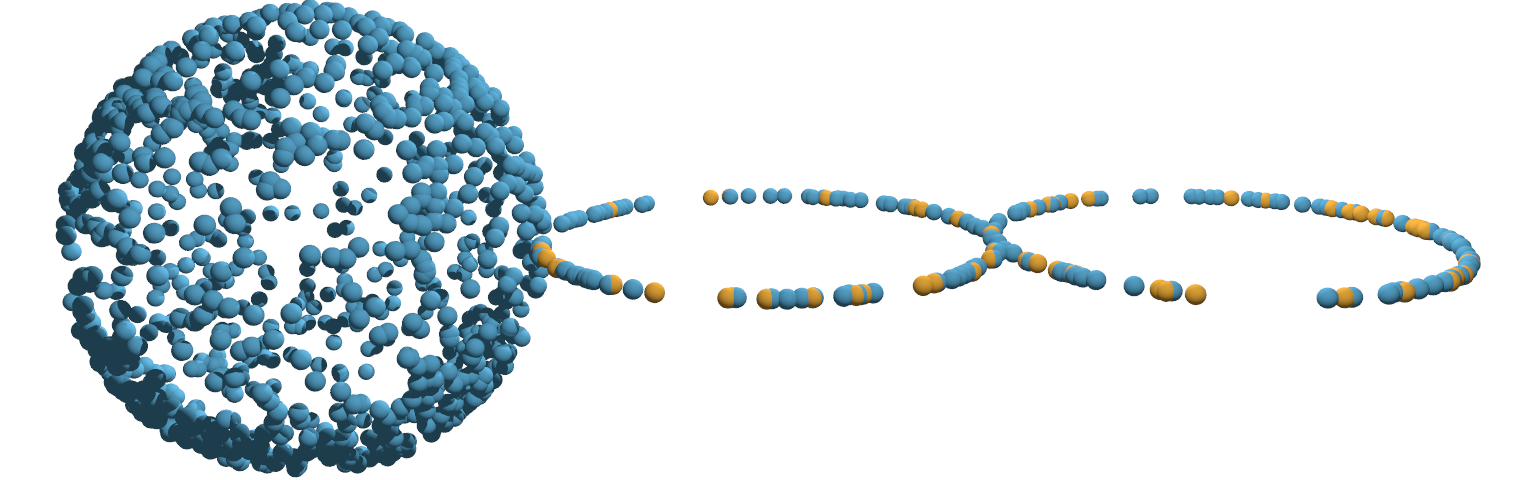}
  \caption{Graph spectral clustering of the vertices of the graph underlying the VR complex of \reffig{retarded-wedge-clustered}.}
  \label{fig:comparespecclust_mickey}   
\end{figure}

\begin{figure}[htbp]
  \vspace*{-0.5cm}
  \centering
  \includegraphics{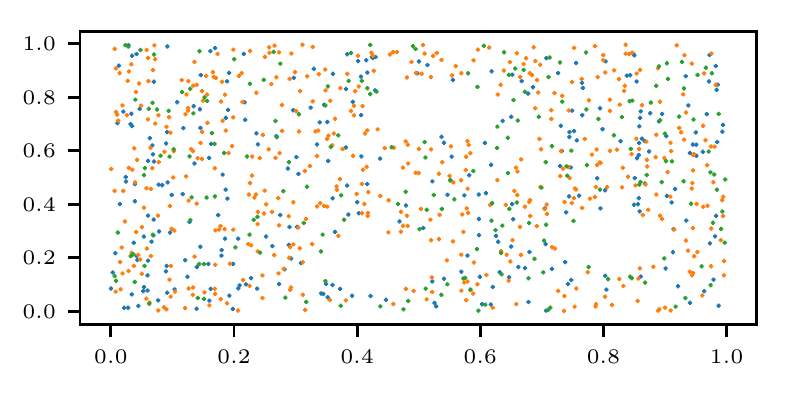}
  \vspace*{-0.5cm} 
  \caption{Graph spectral clustering of the vertices of the graph underlying the alpha complex from \reffig{planeclust}.}
  \label{fig:comparespecclust_punplane}   
\end{figure}

It is worth comparing clustering obtained from our method with the ones obtained by clustering the nodes of the graph underlying each simplicial complex using the graph spectral clustering algorithm. \Reffig{comparespecclust_mickey} shows the results of graph spectral clustering on the nodes of the graph underlying the complex in \reffig{retarded-wedge-clustered}. The two first graph Laplacian eigenvectors were used to map the nodes into $\RR^2$, and then $k$-means was used to find two clusters. Similarly, \reffig{comparespecclust_punplane} displays three clusters on the nodes of the graph underlying the complex representing a punctured plane with three holes in \reffig{planeclust}. The nodes are mapped to $\RR^3$ using the three first graph Laplacian eigenvectors, after which $k$-means was used to find three clusters. In both cases we see that the clusters do not reflect any obviously meaningful property of the underlying data, unlike our method, which clusters in a way sensitive to homology.

\section{Conclusions and future work}
In this paper we have presented a novel clustering method for simplicial complexes, one that is sensitive to the homology of the complex. We see the method as a contribution to an emerging field of spectral TDA~\cite{cascade,barbarossa2018learning}. Our results suggest that the algorithm can be used to extract homological features for simplices of any degree. Experiments in various simplicial complexes demonstrate the ability of the method to accurately detect edges belonging to different non-bounding cycles. Similar results, not shown in this article for practical considerations of visualization, have been obtained by clustering simplices in higher dimensions. The sub-problem of the structure of the linear subspaces in the image of $\psi$, and how to accurately cluster based on demand, require further investigation of both a mathematical and a algorithmic nature.

While our method seems robust to noise in the underlying data, a more thorough investigation into the output's dependence on noise, and the output's dependence on the scale at which a point-cloud-derived simplicial complex is built, is warranted.

Moreover, it has not eluded us that the method as outlined is not restricted to clustering just simplices. Other finitely generated chain complexes, such as discrete Morse complexes or cubical complexes, naturally lend themselves to the same analysis. One may also want to consider if there are theoretical implications even in the smooth case.

Further development will also include enlarging the target of the projection in $\psi$ to include non-zero eigenvectors of $\lap_p$, as in graph spectral clustering. Preliminary results indicate that this yields a further refinement of the homologically sensitive clusters into ``fair'' subdivisions.

Finally, further work needs to explore the effects of weighting. Both structural weighting, \ie/ deriving weights from the local connectivity properties of the complex, as is often done with graph spectral clustering, and weighting originating from the underlying data itself, as is common in TDA.

A potential future application that we suspect fits our method well is collaboration networks~\cite{patania2017}, where $n$-fold collaborations clearly cannot accurately be encoded as $n \choose 2$ pairwise ones.

\section*{Acknowledgments}

Alpha complexes and persistent homology were computed using \textit{GUDHI}~\cite{GUDHI}. Eigenvector computations were done with \textit{SLEPc}~\cite{SLEPc}.

We would like to thank K.\ Hess for valuable discussions.

Both authors were supported by the Swiss National Science Foundation grant number 200021\_172636.

\printbibliography

\end{document}